\documentclass[runningheads,a4paper]{llncs}
\usepackage{amssymb}
\usepackage{graphicx}
\usepackage{booktabs}
\usepackage{tabularx}
\usepackage{float}
\usepackage{url}

\newcommand{\keywords}[1]{\par\addvspace\baselineskip
\noindent\keywordname\enspace\ignorespaces#1}

\begin{document}
\mainmatter  

\title{Real-Time Detection of Electronic Components in Waste Printed Circuit Boards: A Transformer-Based Approach}

\titlerunning{Detection Transformers on Waste PCBs}
%
%
\author{
Muhammad Mohsin\inst{1}
\and Stefano Rovetta\inst{1,2} \\
Francesco Masulli\inst{1,2} 
\and Alberto Cabri\inst{2,3}
}

\institute{
DIBRIS, University of Genoa, 16146, Genoa, Italy
\and Vega Research Laboratories s.r.l., 16121, Genoa, Italy
\and Department of Computer Science, University of Milan, 20133, Milan, Italy
}

\authorrunning{M.Mohsin, S. Rovetta, F. Masulli, A. Cabri}


\maketitle

\begin{abstract}
Critical Raw Materials (CRMs) such as copper, manganese, gallium, and various rare earths have great importance for the electronic industry. To increase the concentration of individual CRMs and thus make their extraction from Waste Printed Circuit Boards (WPCBs) convenient, we have proposed a practical approach that involves selective disassembling of the different types of electronic components from WPCBs using mechatronic systems guided by artificial vision techniques. In this paper we evaluate the real-time accuracy of electronic component detection and localization of the Real-Time DEtection TRansformer model architecture.  Transformers have recently become very popular for the extraordinary results obtained in natural language processing and machine translation. Also in this case, the transformer model achieves very good performances, often superior to those of the latest state of the art object detection and localization models YOLOv8 and YOLOv9.

\end{abstract}

\section{Introduction}

Critical Raw Materials (CRMs), such as copper, manganese, gallium, and various rare earths, are defined by the European Commission as raw materials of great importance to the European Union economy (in particular for the electronic industry) also for the high risk associated with their supply. The list of CRMs is subject to regular review and update by the European Commission \cite{eucrm}.

Waste Printed Circuit Boards (WPCBs) contain several types of electronic components such as integrated circuits (ICs), capacitors, and transistors that have been created using different types and quantities of CRM. Current WPCBs recycling plants are primarily aimed at extract precious metals such as gold and do not provide solution for extracting critical raw materials since their concentration does not allow an effective economic benefit even if greater than in physical mines~\cite{NIU2023137815}.

To increase the concentration of individual CRMs and thus make their extraction from WPCBs more convenient, a practical approach involves the selective disassembling of different types of electronic components from WPCBs using mechatronic systems guided by artificial vision techniques~\cite{cabri2022recovering}.
The selective disassembly of the different types of electronic components present on a WPCBs allows to increase the concentration of specific CRMs and therefore make their extraction more efficient.
To this aim, we proposed the application of electronic component recognition and localization techniques based on Deep Neural Networks to guide a robotic system~\cite{cabri2022recovering,mohsin2024virtual}.
In this paper we will evaluate the real-time accuracy of electronic component detection and localization of the Real-Time DEtection TRansformer (RT-DETR) model architecture~\cite{lv2023detrs}. Transformers have recently become very popular for the exceptional success obtained in natural language processing and machine translation. The performances  of the DEtection TRansformer will also be compared with those of latest state of the art object detection and localization models YOLOv8 and YOLOv9~\cite{mohsin2024rtsi}. Even if we achieved the consolidated results with CNNs and YOLO architectures, the emerging hype of the community towards the transformers gave us the impulse to test the ability of these revolutionary models in the present line of research. Which is why the research task reported in the present paper is the assessment of transformers performance on our developed custom V-PCB dataset with attention to leveraging edge computing devices.

The rest of the paper is structured as follows: Section 2 surveys the state of the art of  WPCBs electronic component detection, Section 3 presents the detailed methodology of the transformer-based object detection model, Section 4 describes the experimental results and their discussion and Section 5 draws the conclusions.

\section{State of the Art}
The electronic components present on the WPCBs have a wide range of sizes, colours, and shapes, which have changed over time due to advancements in PCB manufacturing materials. Due to the mix of recent and old technologies in WPCBs, the electronic component  recognition task is very challenging.

Different studies used object detection models in WPCBs recycling according to their end requirement. For instance, Lu et al.\ \cite{lu2022automatic} used custom dataset of electronic components and implemented the YOLOv3 object detection model to classify different electronic components. They also proposed an automatic sorting system for fast recycling of the detected electronic components. Sharma et al.\ \cite{sharma2024computer} used YOLOv3 model for detecting the electronic components on WPCBs using PCB-DSLR \cite{pramerdorfer2015dataset} and on custom dataset. 

Cabri et al.\ \cite{cabri2022recovering} proposed a YOLOv5-based model on the edge with a custom WPCBs dataset and achieved state of the art results. As a further extension to this work, Mohsin et al.~\cite{mohsin2024measuring} prepared a local environment for data acquisition and developed a robust dataset, named V-PCB, under recycling conditions such as varying light, different camera sources and different view points. To further enhance the WPCBs dataset and for better generalization, they used data augmentation and transfer learning approaches applied with YOLOv5 model. As an additional step,  the recyclability of individual components from the WPCBs was measured to efficiently extract the CRMs. 

\section{Methodology}
In the following section, we will describe the steps performing for the development of our computer vision-based system supporting electronic component disassembly.

\subsection{Dataset Preparation}
A waste PCBs dataset named \textit{V-PCB} has been prepared at Vega Research Laboratories s.r.l. The dataset contains high resolution 747 augmented WPCBs of 65 unique PCBs boards captured under different lighting conditions and different camera sources. The number of electronic components on each board varies depending on the board model. However, the total number of components in the V-PCB dataset is constant.

\begin{table}[ht]
\centering
\caption{Summary of Dataset Characteristics}
\begin{tabular*}{0.6\textwidth}{@{\extracolsep{\fill}} lc}
\toprule
\textbf{Attribute}          & \textbf{Value}           \\
\midrule
Total Images                & 747                      \\
Image Resolution            & 1920x1080 pixels         \\
Dataset Size                & 5 GB                    \\
Annotation Type             & Bounding Boxes           \\
Classes of Components       & 8                        \\
\bottomrule
\end{tabular*}
\label{tab:dataset_characteristics}
\end{table}

Table \ref{tab:dataset_characteristics} shows the summary of the dataset characteristics. Presently, there are eight different classes of components, namely capacitors, coils, diodes, electrolytic capacitors, integrated circuits (ICs), resistors, transformers, and transistors. The dataset is unbalanced for some classes where the presence of such components on the WPCBs is much rarer than other types, therefore making it difficult to collect image samples. Nevertheless, we are still collecting new image samples to equally represent all classes.

\subsection{Model Architecture}
\begin{figure}
\centering
\includegraphics[width=12.5cm, height=6.5cm]{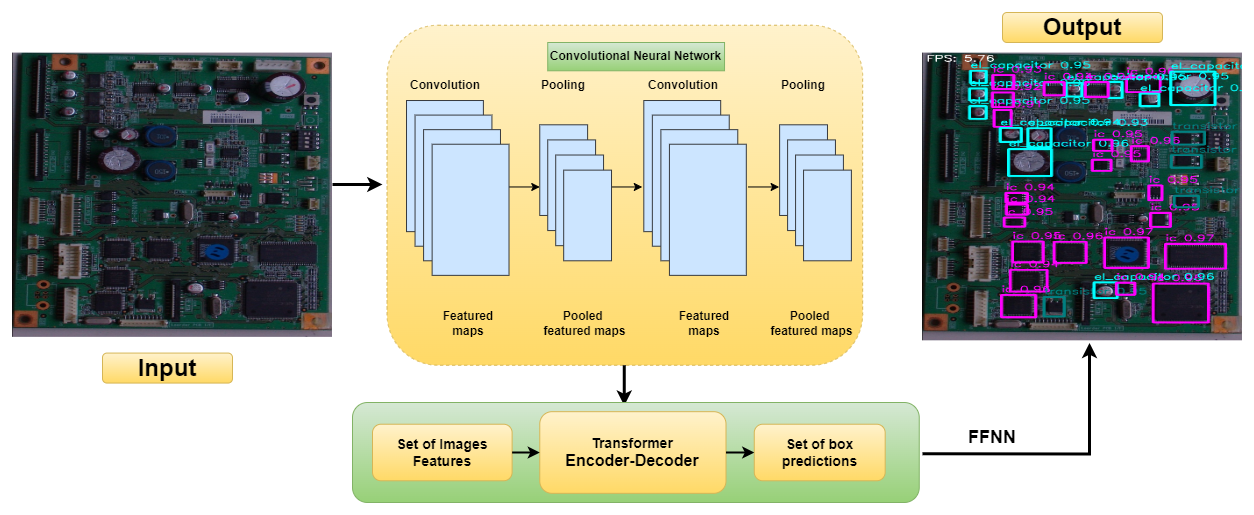}
\caption{Block diagram of detection transformer for WPCBs component detection}
\label{fig:detr}
\end{figure}

DEtection TRansformer (DETR) is an innovative algorithm for object detection proposed by researchers of Facebook AI Research in 2020 \cite{carion2020end}. The model employs the transformer architecture, a highly efficient sequence-to-sequence model widely used in various natural language processing tasks. Figure \ref{fig:detr} shows the block diagram of detection transformer architecture for WPCBs component detection.
DETR is an object detector that uses a convolutional backbone, with a Transformer on top, to process sets of data. The system employs a conventional Convolutional Neural Network (CNN) architecture to obtain a two-dimensional representation of an input image. Prior to feeding it into a transformer encoder, the model compresses the input and improves it by adding positional encoding. A transformer decoder takes just a small number of pre-learned positional embeddings, referred to as object queries, and also focuses on the output of the encoder. Each output embedding of the decoder is fed into a common feed forward neural network (FFNN) which makes predictions about either a detection (class and bounding box) or a \textit{no-object} class. The vision transformer architecture consists of three components: 
\begin{enumerate}
    \item A CNN Backbone, whose primary objective is to extract the features or attributes of an image.
    \item The encoder and decoder models, which receive the features as input and transforms the positional embedding to feed the FFNN
    \item A FFNN that makes predictions about the identified classes and their bounding boxes.
\end{enumerate}

In Real-Time DEtection TRansformer (RT-DETR)~\cite{lv2023detrs}, encoder takes input in final three stages of the backbone, specifically S3, S4, and S5. The hybrid encoder efficiently converts multi-scale data into a sequence of image features by utilising the Attention-based Intra-scale Feature Interaction (AIFI) and CNN-based Cross-scale Feature Fusion (CCFF). The IoU-aware query selection is then employed to determine a precise quantity of image features that will serve as the initial object queries for the decoder. In the end, the decoder utilises additional prediction heads to systematically improve object queries in order to generate bounding boxes and confidence ratings.

\subsection{Training Procedure and Evaluation Metrics}
Object detection models come in various sizes to meet various requirements and computational resources. For example, there are large, and extra-large models. The large RT-DETR model requires more computational resources, Floating Point Operations (FLOPs) and memory during the training phase compared with extra-large models, resulting in faster inference times and making it suitable to run on edge computing devices, whereas the extra-large RT-DETR models require more power, memory space and are developed specifically for tasks that need high precision and complex visualisation of objects. The selection amongst them depends upon the particular requirements of the task, the available computational resources, and the choices that must be made regarding accuracy and performance.
In this work we used both large and extra-large RT-DETR models of detection on the V-PCB dataset, to asses the real-time performance of detection system as well as accuracy of the detected objects.

The models are trained on 300 epochs with two different batch sizes, 4 and 8. The model training started by employing a transfer learning technique, leveraging the detection transformer model weights as the base model. Following that, the training focused on improving the model, specifically using the WPCBs dataset. During this stage, the model weights are adjusted to optimize the detection accuracy, particularly for high-resolution PCB images. The technique of fine-tuning significantly enhances the model's ability to precisely identify and localize different components on WPCBs in real-time.

For evaluating the models performance on real-time environment, we use the Mean Average Precision (mAP) metrics. It is a widely used metric for object detection's models evaluation.
The mean Average Precision (mAP) is computed by taking the average of the Precision-Recall curve over various Intersection over Union (IoU) thresholds and object classes \cite{padilla2021electronics}.

\section{Experimental Results and Discussion}

The experiments were held on a  Linux Ubuntu server platform equipped
with 2 Intel Xeon Silver 4214 CPUs running at 2.20GHz, 93GB of
RAM, and an NVIDIA Quadro RTX5000 with 16GB, 3072 CUDA
(Compute Unified Device Architecture) cores, 384 Tensor cores,
and 48 RT (ray tracing) cores on the GPU (Graphics Processing
Unit). The programming language used is Python 3.8.

Table \ref{tab:overall_performance} shows the training details and the performance indexes  both large and extra-large RT-DETR models. To compare the performance of both models, the data split ratio, number of epochs and batch size are kept the same. The training time and memory usage of the large model are lower compared to the extra-large model, however this comes at the expense of a worse total mAP, particularly mAP50-95.  It is clear that a large model is more efficient based on its lower complexity and lower memory utilization. We evaluated the model performance on an average threshold of IoU 50\%, as illustrated by mAP50. Further, the average of mAP across multiple IoU thresholds (mAP50-90), ranging from 50\% to 95\%, is computed to obtain more detailed results. The results of both models are presented in Table \ref{tab:overall_performance}. Conversely, the larger models showed excellent performance but required more memory and training time. This indicates that the selection of a model is entirely dependent upon the available computational resources and requirement for accuracy. For instance, in industrial applications, edge systems are used to detect objects real-time and require low computational resources. Drone systems, for example, rely on edge computing for real-time object detection. Medical imaging tasks, however, require more powerful hardware to provide accurate diagnoses.  
\begin{table}
\centering
\caption{Training details and  performance indexes of large and extra-large RT-DETR models}
\label{tab:overall_performance}
\setlength{\tabcolsep}{1em} 
\begin{tabularx}{0.67\textwidth}{@{}lcc@{}}
\toprule
\textbf{Parameter} & \textbf{L-Model} & \textbf{XL-Model} \\
\midrule
Epochs & 300 & 300 \\
Training Time (hours) & 6.85 & 24.45 \\
Images & 65 & 65 \\
Instances & 3682 & 3682 \\
Overall Precision (P) & 0.97 & 0.98 \\
Overall Recall (R) & 0.98 & 0.98 \\
\textbf{Overall mAP50} & \textbf{0.98} & \textbf{0.99} \\
\textbf{Overall mAP50-95} & \textbf{0.82} & \textbf{0.84}\\
\bottomrule 
\end{tabularx}
\end{table}

\begin{figure}[!htb]
\centering
\includegraphics[width=.75\textwidth]{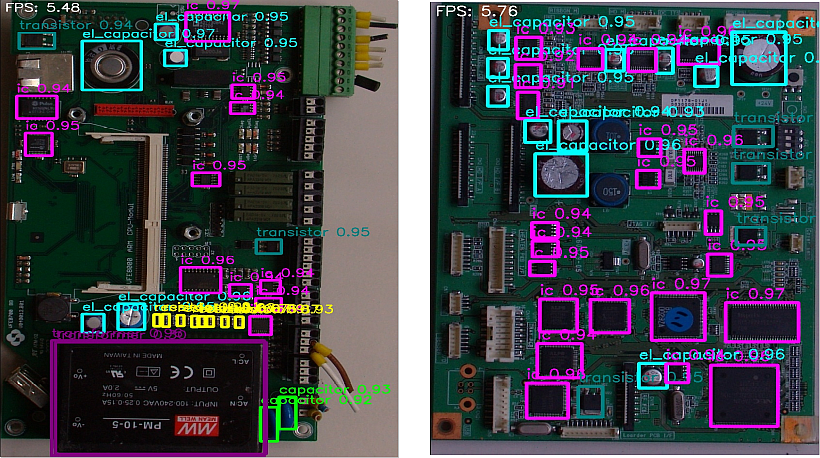}
\caption{Some results of electronic component detection}
\label{fig:final_results}
\end{figure}

Precision, recall, and mean average precision at the IoU threshold of 50\% and 50-90\% are calculated during the training process. The results obtained show the model's performance gradually improves as the number of epochs increases. This linear increase in performance shows that the transformer-based object detection models generalisation capability on the custom V-PCB dataset. Performance is evaluated by using two different batch sizes, namely 4 and 8. Fig.\ref{fig:final_results} shows some results of the RT-DETR object detection model in detecting and localizing the electronic components on WPCBs.
\begin{table}[!htb]
\centering
\caption{Comparison of RT-DETR and YOLO models performance on V-PCB dataset}
\label{tab:performance_comparison}
\setlength{\tabcolsep}{1em} 
\begin{tabularx}{0.85\textwidth}{@{}lcccc@{}}
\toprule
\textbf{Model} & \textbf{F1 Score} & \textbf{mAP50} & \textbf{mAP50-95} & \textbf{Latency (ms)} \\
\midrule
YOLOv8  & 0.88 & 0.90 & 0.82 & 26 \\
YOLOv9  & 0.90 & 0.91 & 0.85 & 21 \\
\textbf{RT-DETR}& \textbf{0.98} & \textbf{0.99} & \textbf{0.84} & \textbf{20} \\
\bottomrule
\end{tabularx}
\end{table}

Table \ref{tab:performance_comparison} shows the comparison of the transformer-based object detection \cite{terven2023comprehensive} model RT-DETR with latest state of the art models YOLOv8 and YOLOv9 on V-PCB dataset \cite{mohsin2024rtsi}. RT-DETR outperforms both YOLO models concerning F1 Score, mAP50 and detection speed in real-time (latency), while concerning mAP50-95 YOLOv9 is little better than RT-DETR.

\section{Conclusions}
Motivated by the exceptional success of Transformers in natural language processing and machine translation, in this paper we evaluated the application of the Real-Time DEtection TRansformer model~\cite{lv2023detrs}, to the recognition and localization of electronic components on WPCB to further enhance the disassembly and sorting process of electronic components in a circular economy perspective aimed at extracting CMR from electronic waste.

Further, the effectiveness of feature extraction and the ability of scaling and flexibility of the model have been explored, evaluating the model performance using a custom dataset developed under recycling environments at Vega Research Laboratories s.r.l.

The transformer model obtained very good performances, often superior to those of the latest state of the art object detection and localization models YOLOv8 and YOLOv9.

\section{Acknowledgements}
The project is partially funded by EIT Raw Materials grant number 15099-sclc-2020-8. The Ph.D. position of the first author is funded by MUR (DM n. 360 of 04-21-2022).

\bibliographystyle{plain} 
\bibliography{ref}
\end{document}